\pdfoutput=1

\documentclass[11pt]{article}

\usepackage{acl}

\usepackage{booktabs}
\usepackage[inline]{enumitem}
\usepackage[skip=3pt]{caption}
\usepackage{acronym}
\usepackage{multirow}
\usepackage{makecell}
\usepackage{graphicx}
\usepackage{xcolor}
\usepackage{subcaption}
\usepackage{wrapfig}
\usepackage{xspace}

\usepackage[T1]{fontenc}
\usepackage[utf8]{inputenc}
\usepackage{microtype}

\usepackage{times}
\usepackage{latexsym}

\usepackage[utf8]{inputenc}
\usepackage{tcolorbox}

\usepackage[T1]{fontenc}

\usepackage[utf8]{inputenc}
\usepackage{amsmath}
\usepackage[ruled,vlined]{algorithm2e}

\usepackage{microtype}

\usepackage{inconsolata}

\usepackage{graphicx}

\usepackage{algpseudocode}

\usepackage{tikz}
\usetikzlibrary{shapes.geometric, arrows, positioning, fit, calc, patterns}
\usepackage{graphicx}

\usepackage{tikz}
\usepackage{pgfplots}

\pgfplotsset{compat=1.18} %
\usepackage{pgfplotstable} %

\usepackage{subcaption}

\usepackage{caption}

\usepackage{xcolor}

\usetikzlibrary{patterns,positioning,calc}

\usepackage{newfloat}
\DeclareFloatingEnvironment[fileext=lop, listname={List of prompts}, 
                            name=Prompt, placement=h]{prompt}

\usepackage{listings}
\usepackage{xcolor}
\usepackage{listings}
\usepackage{url}
\usepackage{soul}
\title{Learning to Judge: LLMs Designing and Applying Evaluation Rubrics}

\author{
  \textbf{Clemencia Siro\textsuperscript{1}},
  \textbf{Pourya Aliannejadi\textsuperscript{2}},
  \textbf{Mohammad Aliannejadi\textsuperscript{3}}
\\
\\
  \textsuperscript{1}Centrum Wiskunde \& Informatica (CWI), Amsterdam, The Netherlands \\
  \textsuperscript{2}Shahid Beheshti University, Tehran, Iran \\
  \textsuperscript{3}University of Amsterdam, Amsterdam, The Netherlands
\\
\\
  \small{ 
    \{
    \texttt{c.n.siro@cwi.nl},
    \texttt{alianpourya@gmail.com},
    \texttt{m.aliannejadi@uva.nl}\}
  }
}

\begin{document}
\maketitle
\begin{abstract}
Large language models (LLMs) are increasingly used as evaluators for natural language generation, applying human-defined rubrics to assess system outputs.
However, human rubrics are often static and misaligned with how models internally represent language quality.
We introduce \textbf{GER-Eval} (\textit{Generating Evaluation Rubrics for Evaluation}) to investigate whether LLMs can \textit{design} and \textit{apply} their own evaluation rubrics.
We evaluate the semantic coherence and scoring reliability of LLM-defined criteria and their alignment with human criteria.
LLMs reliably generate interpretable and task-aware evaluation dimensions and apply them within models, but their scoring reliability degrades in factual and knowledge-intensive settings.
Closed-source models such as GPT-4o achieve higher agreement and cross-model generalization than open-weight models such as Llama.
Our findings position evaluation as a learned linguistic capability of LLMs, consistent within models but fragmented across them, and call for new methods that jointly model human and LLM evaluative language to improve reliability and interpretability. Our code and generated rubrics can be found here.~\footnote{\url{https://github.com/Clemenciah/llm-generated-rubrics}}

\end{abstract}

\section{Introduction}

Recent advances in large language models (LLMs) have shifted evaluation of generated text in various NLP and IR tasks beyond traditional surface metrics like BLEU and ROUGE toward model-based judgments that better capture fluency, coherence, and informativeness~\cite{dubois2023alpacamatch,zheng2023judging}. LLMs are increasingly used as evaluators by applying natural language rubrics or preference prompts to approximate human judgments across diverse NLG tasks~\cite{chiang2024selfeval, dubois2023alpacamatch, zheng2023judging}. However, this raises important questions about the reliability, consistency, and linguistic biases of model judgments. In particular, it remains unclear whether LLMs evaluate text using the same dimensions and cues intended by humans when defining evaluation rubrics, or whether their assessments are shaped by internal representational biases.

Recent studies show that ``LLM-as-judge'' suffers from both \textit{systemic biases} and \textit{limited adaptability}. LLM-based evaluations are sensitive to prompt phrasing \cite{zhou2023promptbench}, label order \cite{wang2023fair}, response length \cite{stureborg2024inconsistent}, and position \cite{shi2024positionbias,kudugunta2023preferencemodel}. Studies have also shown that LLMs favor their own generations \cite{koo2023cobbler}, or fail to produce stable preferences across reruns~\citep{stureborg2024inconsistent}. While prompting tweaks and calibration strategies have been proposed \cite{jang2023correctbias, kudugunta2023preferencemodel}, these methods often address superficial variances rather than the deeper challenge, i.e., LLMs are instructed to interpret static human-defined rubrics that may not align with their internal language quality representation.

A growing body of work explores whether LLMs can take a more active role in the evaluation process, moving beyond simply applying predefined rubrics. Prior work explores LLMs producing reward signals during training \cite{yuan2024selfreward}, decomposing evaluation into human-aligned fine-grained criteria \cite{liu2024hdeval}, and collaborating interactively to define evaluation functions \cite{shankar2024evalgen}. The BSM framework \cite{saha2024bsm} prompts LLMs to generate task-specific criteria, applies them, and aggregates judgments for output comparison. 
Existing research builds on the assumption that the LLMs' self-defined rubrics are reliable and lead to improved and consistent scores, as opposed to human-defined rubrics. However, this assumption has not yet been systematically tested and verified in the literature.

To investigate this alignment between model- and human-defined evaluation rubrics, we propose \textbf{GER-Eval} (\textit{Generating Evaluation Rubrics for Evaluation}), a unified framework for studying whether LLMs can autonomously \textit{generate} and \textit{apply} their own evaluation rubrics. Importantly, GER-Eval is designed as a diagnostic framework. Rather than introducing a new rubric generation method, our goal is to systematically test a core assumption underlying recent LLM-as-judge approaches: LLM-generated rubrics are more reliable and can be applied consistently.
By decoupling rubric generation from rubric application, we enable controlled analyses of \emph{where} this assumption holds and \emph{where} it fails across tasks, prompt conditions, and model families. 
GER-Eval is a two-stage evaluation framework that decouples \textit{rubric design and generation} from \textit{rubric application}. During generation, LLMs generate task-specific evaluation criteria, including rubric names, definitions, and scoring scales. In the subsequent application stage, these self-generated rubrics are used by LLMs to score candidate outputs, either in a zero-shot setting or augmented with few-shot demonstrations.

Our experiments across four datasets on NLG tasks show that LLMs can generate coherent and interpretable rubrics that capture task-relevant quality dimensions such as coherence, fluency, and informativeness, showing consistency across prompting conditions.  
These rubrics enable consistent scoring within models but exhibit limited cross-model agreement and weaker alignment with human judgments, suggesting that evaluation rubrics are model-dependent.
Closed-source models like GPT-4o achieve higher internal consistency and stronger alignment with human judgments than open-weight models such as Llama. In contrast, all models perform less reliably in factual or knowledge-intensive settings.

\noindent \textbf{Contributions.}
We introduce \textbf{GER-Eval}, a unified framework for analyzing how LLMs \textit{generate}, \textit{apply}, and \textit{generalize} their own evaluation rubrics.
Through large-scale experiments across multiple NLG tasks and models, we assess rubric coherence, scoring reliability, and human alignment.
Our findings reveal systematic differences between closed-source and open-weight models, advancing our understanding of LLMs as autonomous evaluators.

\section{Related Work}

Evaluation in NLP has traditionally relied on human-annotated benchmarks and predefined rubrics tailored to specific tasks. For example, SummEval \cite{fabbri-etal-2021-summeval} and QAGS \cite{zhong-etal-2020-qags} use human-authored criteria such as coherence, relevance, and factuality to assess the quality of summarization outputs. While these methods offer reliable gold standards, they are expensive to scale and often lack flexibility across domains and tasks.
With the rise of LLMs, recent work has explored using LLMs as automated evaluators. Benchmarks such as AlpacaEval \cite{dubois2023alpacamatch}, MT-Bench \cite{zheng2023judging}, and HelpSteer \cite{xu2023steering} prompt LLMs to rate or rank responses according to human-defined rubrics. These approaches enable scalable evaluation, but inherit the limitations of the rubrics they rely on and are prone to systemic biases. G-Eval \cite{liu2023geval} improves the robustness of LLM-based evaluation through prompt refinement, while Fu et al. \cite{fu2023debiasing} focus on mitigating common biases such as verbosity and position effects. PAIRS \cite{liu2024pairs} reframes evaluation as a ranking task and proposes an uncertainty-guided pairwise comparison method that achieves more stable and human-aligned output rankings.
A complementary line of work considers whether LLMs can take a more active role in the evaluation process. HD-Eval \cite{liu2024hdeval} introduces a hierarchical decomposition framework that iteratively generates and refines fine-grained criteria aligned with human preferences using attribution-based pruning and aggregation. EvalGen \cite{shankar2024evalgen} presents a mixed-initiative interface in which LLMs and users co-construct evaluation functions. Their study reveals that evaluation criteria often emerge through interaction with model outputs, a phenomenon they describe as \textit{``criteria drift''}. The BSM framework \cite{saha2024bsm} advances this direction by prompting LLMs to generate task-specific evaluation criteria, apply them to paired outputs, and aggregate the results through a modular evaluation pipeline.

Existing research builds on the assumption that the LLMs' self-defined rubrics are reliable and lead to improved and consistent scores, as opposed to human-defined rubrics. This assumption has not yet been systematically tested and verified in the literature.
It remains unclear how reliably LLMs can generate and apply self-defined rubrics in ways that lead to consistent scoring behavior and meaningful alignment with human evaluation criteria. Therefore, we propose \textbf{GER-Eval} to investigate whether LLMs can generate and apply their self-defined rubric and do they align with human rubric.

\section{GER-Eval: A Framework for LLM-Generated Evaluation Rubrics}\label{sec:gemeval}

We propose \textbf{GER-Eval} (\textit{Generating Evaluation Rubrics for Evaluation}), a framework to investigate whether LLMs can \textit{generate} and \textit{apply} their self-defined rubrics. GER-Eval is a two-stage framework that decouples \textit{rubric generation} from \textit{rubric application}, enabling us to analyze two complementary questions: (i) How LLMs conceptualize evaluation rubrics?; and (ii) How consistently do LLMs apply their own evaluation rubrics?

\subsection{Notation}  
Let $p_\theta$ denote an LLM with parameters $\theta$. Given a task description $t$, and prompting condition $\pi$,
 the model generates an evaluation criterion $M = \{m_1, \ldots, m_k\}$, where each criterion is represented as a triple $m_i = (n_i, d_i, s_i)$ with a \textit{name}, \textit{description}, and \textit{scale} (numeric or categorical). Each criterion also includes a short instruction $I(m_i)$. Candidate outputs to be evaluated are denoted by $Y = \{y_1, \ldots, y_N\}$.  

\subsection{Rubric Generation}  
The evaluation rubric generation stage conditions the model on a prompt $prompt_{gen}(t,\pi)$, where $\pi$ specifies the \textbf{prompting condition}. The model then generates a set of criteria:  
\[
M \sim p_\theta(M \mid prompt_{gen}(t,\pi))~.
\]
We experimented with three prompting conditions:  \textit{Task-only}, \textit{Task + Contexts}, and \textit{Task + Contrastive Examples}.
We include these three conditions to examine two aspects: (i) whether LLMs generate consistent rubrics across prompt variations, which reflects the stability of their task understanding; and (ii) whether additional grounding through demonstrations leads to more specific and domain-sensitive criteria.

For each criterion $m_i$, the model also generates a short instruction $I(m_i)$ describing how the criterion should be scored:  
\[
I(m_i) \sim p_\theta(I \mid m_i, t)~.
\]

\subsection{Rubric Application}  
In the application stage, the generated criteria are used to assess candidate outputs. Given a criterion $m_i$ with instruction $I(m_i)$ and an output $y_j \in Y$, the model is conditioned on a scoring prompt 
$prompt_{eval}$
$(t, m_i, I(m_i), y_j, \pi)$. 
$prompt_{eval}$
includes the task description, the criterion (name/description/scale), its instruction, and the candidate output, with $\pi$ controlling whether demonstrations are included, to generate both reasoning $r_{i,j}$ and a score $s_{i,j}$:

\begin{equation*}
\begin{aligned}
\pi^{eval}_{i,j} &= 
prompt_{eval}\!\bigl(t, m_i, I(m_i), y_j, \pi\bigr), \\
(r_{i,j}, s_{i,j}) &\sim p_\theta\!\bigl(r, s \mid \pi^{eval}_{i,j}\bigr).
\end{aligned}
\end{equation*}

Scores may be numerical (e.g., 1--5) or categorical (e.g., \textit{high/medium/low}), depending on the scale $s_i$.  

We test two prompting conditions $\pi$: \textit{Rubric-only (zero-shot)} and \textit{Rubric + Demonstrations (few-shot)}
to evaluate whether LLMs can reliably apply their own rubrics with or without demonstrations, and compare their alignment with human rubrics

\begin{algorithm}[t]
\caption{GER-Eval Framework}
\label{alg:gem-eval}
\DontPrintSemicolon
\SetAlgoNlRelativeSize{-1}
\SetAlgoNlRelativeSize{-1}
\SetAlgoNlRelativeSize{-1}
\KwIn{Task description $t$, prompting condition $\pi$, outputs $Y$}
\KwOut{Structured evaluation results}
$M \sim p_\theta(M \mid prompt_{gen}(t,\pi))$ \;
\ForEach{$m_i \in M$}{
  $I(m_i) \sim p_\theta(I \mid m_i, t)$ \;
  \ForEach{$y_j \in Y$}{
    $(r_{i,j}, s_{i,j}) \sim p_\theta(r,s \mid prompt_{eval}(t,m_i,I(m_i),y_j,\pi))$ \;
  }
}
\Return $\{(m_i, I(m_i), \{(y_j, r_{i,j}, s_{i,j})\})\}$ \;
\end{algorithm}

Algorithm~\ref{alg:gem-eval} summarizes the GER-Eval process. GER-Eval is positioned as a framework for studying LLMs as both \textit{designers} and \textit{users} of evaluation criteria and can be broadly applied across tasks and models, and is directly comparable to human-defined evaluation rubrics.

\subsection{Experimental Setup}

To assess GER-Eval across diverse tasks and model families, we design experiments that vary along three dimensions: the datasets under evaluation, the prompting method used in rubric generation and application, and the LLMs employed. 

\subsubsection{Tasks and datasets}  
We evaluate GER-Eval on three NLG tasks: conversational, summarization, and instruction-following, across four established benchmarks: USR~\cite{mehri-eskenazi-2020-usr}, SummEval~\cite{fabbri-etal-2021-summeval}, SumPubMed~\cite{gupta-etal-2021-sumpubmed}, and HelpSteer2~\cite{wang2024helpsteer2}.  Each dataset provides human scores on multiple quality attributes, enabling direct validity analysis via correlation between LLM-based scores and human judgments. Together, they span diverse domains and varying dependence on factual correctness, allowing us to test both generality and failure modes of rubric generation and application. Detailed dataset specifics in Appendix~\ref{app:datasets}.

For each dataset, we extracted the core task description from its corresponding official paper and used it to construct the generation description for the rubric generation stage and the scoring description for the scoring stage.

\subsubsection{Generation prompts}  
For rubric generation, we experimented with three prompting conditions. We consider three types: (i) \textbf{Task-only}, where the model sees only the task description; (ii) \textbf{Task + Contexts}, where representative input contexts (e.g., a news article or dialogue history) are added; and (iii) \textbf{Task + Contrastive Examples}, where contexts are paired with one positive and one negative output. We treat high-scoring outputs as positives and low-scoring outputs as negatives using human ratings. 

\subsubsection{Scoring conditions}  
For evaluation rubric application, we compare two scoring setups: (i) \textbf{Rubric-only (zero-shot)}, where the model applies the rubric without demonstrations; and (ii) \textbf{Rubric + Demonstrations (few-shot)}, where the rubric is presented together with sample outputs and scores.

\subsubsection{Rubric source}  
We evaluate two types of rubrics, both applied by LLMs. In the first case, we use \textbf{human-defined rubrics}, taken directly from the human rubric items and instructions specified in each dataset. In the second, we use \textbf{LLM-generated rubrics}, generated in the evaluation rubric generation stage of GER-Eval. Comparing these two conditions allows us to assess how LLMs apply evaluation criteria originating from humans versus self-defined.

\subsubsection{Models}  
We evaluate both closed- and open-source LLMs to capture a range of capabilities: GPT-4o~\citep{gpt4-report} and GPT-4o-mini~\citep{gpt4-report}, Mixtral-8x22B~\citep{jiang2024mixtralexperts}, Llama-3.3-70B~\citep{Touvron2023-llama}, and Qwen2.5-72B~\citep{qwen2}. This selection reflects models widely adopted in evaluation research and spans different architectures and parameter scales. Details of exact models, model configurations, and implementation details are in Appendix~\ref{app:models}.

\begin{figure}[t]
\centering
\includegraphics[width=1\columnwidth]{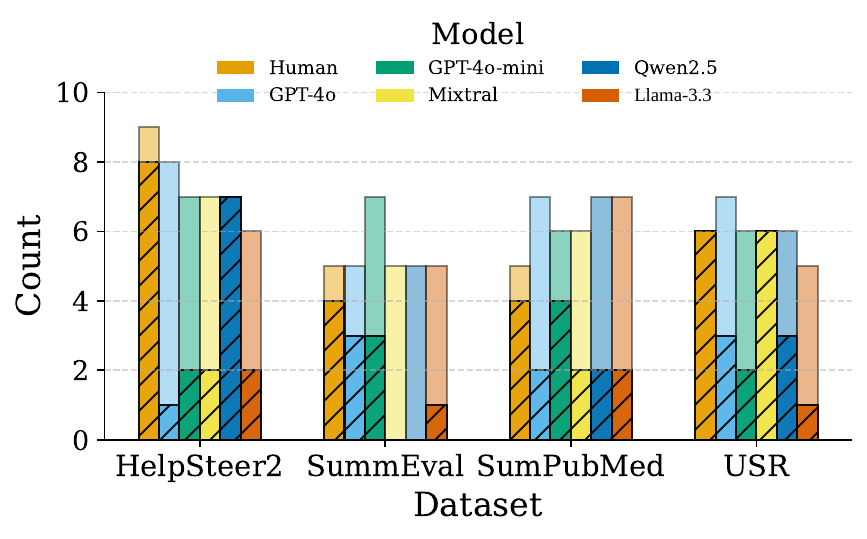}
\caption{
Task-specific counts by model across datasets. 
Each color represents a model; lighter bars denote \textbf{Total} counts, and darker hatched bars denote \textbf{Task-specific} rubrics. 
}
\label{fig:rq2_task_sensitivity_grouped}
\end{figure}

\begin{figure}[!t]
    \centering
  
        \centering
        \includegraphics[width=1\columnwidth]{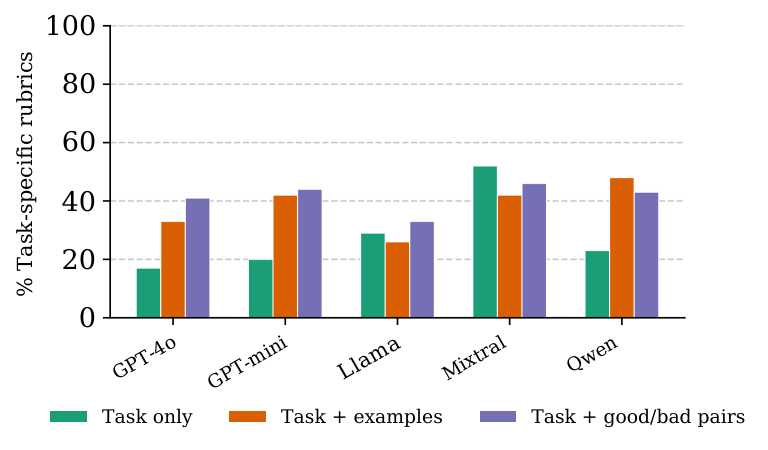}
    
    \caption{Percentage of task-specific rubrics identified by each model under three prompting conditions.}
    \label{fig:task-specificity}
\end{figure}

\section{Results}
In this section, we report our empirical analysis of GER-Eval.

\subsection{Rubric Generation}
\textit{How effectively can LLMs generate explicit, task-specific evaluation rubrics that reflect human-defined quality dimensions?}

\begin{table*}[!t]
\centering
\small
\begin{tabular}{llcccccc}
\toprule
Dataset & Model & \multicolumn{3}{c}{Zero-shot} & \multicolumn{3}{c}{Few-shot} \\
\cmidrule(lr){3-5} \cmidrule(lr){6-8}
 &  & Breadth & Unique (\%) & Align. (\%) & Breadth & Unique (\%) & Align. (\%) \\
\midrule
\multirow{5}{*}{HelpSteer2}
  & GPT-4o & 6 & 100 & 83 & 8 & 88 & 100 \\
  & GPT-4o-mini & 6 & 100 & 67 & 7 & 86 & 67 \\
  & Llama & 5 & 80 & 100 & 6 & 83 & 100 \\
  & Mixtral & 6 & 83 & 100 & 7 & 57 & 75 \\
  & Qwen & 6 & 100 & 83 & 7 & 86 & 100 \\
\midrule
\multirow{5}{*}{SummEval}
  & GPT-4o & 6 & 83 & 80 & 5 & 80 & 75 \\
  & GPT-4o-mini & 6 & 83 & 40 & 7 & 71 & 20 \\
  & Llama & 5 & 60 & 67 & 5 & 60 & 67 \\
  & Mixtral & 5 & 60 & 67 & 5 & 60 & 33 \\
  & Qwen & 5 & 80 & 50 & 5 & 80 & 50 \\
  
\midrule
\multirow{5}{*}{SumPubMed}
  & GPT-4o & 6 & 83 & 100 & 7 & 71 & 60 \\
  & GPT-4o-mini & 7 & 86 & 83 & 6 & 100 & 67 \\
  & Llama & 6 & 33 & 50 & 7 & 86 & 33 \\
  & Mixtral & 5 & 80 & 100 & 6 & 67 & 75 \\
  & Qwen & 6 & 50 & 33 & 7 & 57 & 75 \\
  
\midrule

\multirow{5}{*}{USR}
  & GPT-4o & 5 & 100 & 80 & 7 & 86 & 83 \\
  & GPT-4o-mini & 6 & 100 & 100 & 6 & 100 & 83 \\
  & Llama & 5 & 80 & 75 & 5 & 80 & 100 \\
  & Mixtral & 5 & 80 & 100 & 6 & 83 & 60 \\
  & Qwen & 5 & 100 & 100 & 6 & 100 & 83 \\
  
\bottomrule
\end{tabular}
\caption{Rubric generation results by model and prompting condition. Breadth = raw number of generated rubrics, Unique = percent not repeated, Align. = percentage of unique criteria that map to human rubric criteria. }
\label{tab:criteria-analysis}
\end{table*}

Table~\ref{tab:criteria-analysis} and Table~\ref{tab:model-condition} summarize how LLMs generate evaluation rubrics across datasets and prompting conditions. 
Overall, the models generated five to eight rubrics per dataset, showing an ability to construct structured rubrics. Few-shot prompting generally increases the number of rubrics, while GPT-4o and GPT-4o-mini generate more unique rubrics (around 90\%), indicating broader yet non-redundant coverage. The high uniqueness is matched by strong alignment with human rubrics, where \textbf{Align. (\%)} measures the proportion of generated criteria that can be mapped to a human-defined rubric (Table~\ref{tab:criteria-analysis}), exceeding 80\% across most datasets.

Variation across datasets reflects differences in the focus of the task. On HelpSteer2, where quality involves tone and empathy, several models reach full alignment (i.e., Align. (\%) $=100$ in Table~\ref{tab:criteria-analysis}) in few-shot settings, suggesting that interpersonal aspects are easier to internalize. In contrast, SumPubMed shows the lowest alignment (often below 60\%), as reproducing biomedical terminology and coverage requires specialized knowledge.  SummEval and USR yield moderate to high alignment: SummEval yields moderate alignment due to its abstract linguistic rubrics. USR achieves both high uniqueness and agreement, indicating that conversational settings are easier for models to approximate the human rubric.

Prompting conditions further influence rubric quality. As shown in Table~\ref{tab:model-condition}, contextual and contrastive examples lead to richer and more varied rubrics but affect alignment differently across models. Mixtral attains the highest alignment under task-only prompting (92\%), while Llama improves most with contrastive examples (94\%). GPT-4o remains robust across all setups, showing stable alignment and low rubrics redundancy.

Figures~\ref{fig:rq2_task_sensitivity_grouped} and~\ref{fig:task-specificity} show that prompting and task domain influence task-specificity of LLM-generated rubrics. 
With demonstrations, models generate more task-grounded criteria that reflect domain goals, suggesting that examples help calibrate models' internal quality judgment.  
This effect is stronger in conversational and instruction-following datasets, where models express socially oriented dimensions such as empathy and politeness, but weaker in summarization tasks, where models generate more generic rubrics.  
Across models, GPT-4o and Mixtral generate more task-specific rubrics compared to Llama or Qwen, indicating that both prompt design and model priors influence how evaluation criteria are generated and their alignment with human-defined dimensions.

\begin{table*}[!t]
\centering
\small
\begin{tabular}{lccc|ccc|ccc}
\toprule
 & \multicolumn{3}{c|}{Task-only} & \multicolumn{3}{c|}{Contexts} & \multicolumn{3}{c}{Good/Bad} \\
\cmidrule(lr){2-4}\cmidrule(lr){5-7}\cmidrule(lr){8-10}
Model & Breadth & Unique & Align. & Breadth & Unique & Align. & Breadth & Unique & Align. \\
\midrule
GPT-4o & 23(5.8) & 92 & 86 & 27(6.8) & 81 & 80 & 27(6.8) & 81 & 68 \\
GPT-4o-mini & 25(6.2) & 92 & 72 & 26(6.5) & 89 & 59 & 25(6.2) & 82 & 76 \\
Llama & 21(5.2) & 63 & 73 & 23(5.8) & 77 & 75 & 18(4.5) & 89 & 94 \\
Mixtral & 21(5.2) & 76 & 92 & 24(6.0) & 67 & 61 & 26(6.5) & 62 & 82 \\
Qwen & 22(5.5) & 82 & 67 & 25(6.2) & 81 & 77 & 23(5.8) & 86 & 65 \\
\bottomrule
\end{tabular}
\caption{Rubric generation results by model and prompting condition. Breadth = total number of rubrics across datasets with average per dataset in parentheses (e.g., 23(5.8)), Unique = average \% not repeated rubrics, Align. = average \% overlapping with human rubrics.}
\label{tab:model-condition}
\end{table*}

\begin{figure*}[!t]
    \centering
    \subfloat[USR dataset]{%
        \includegraphics[width=0.7\columnwidth]{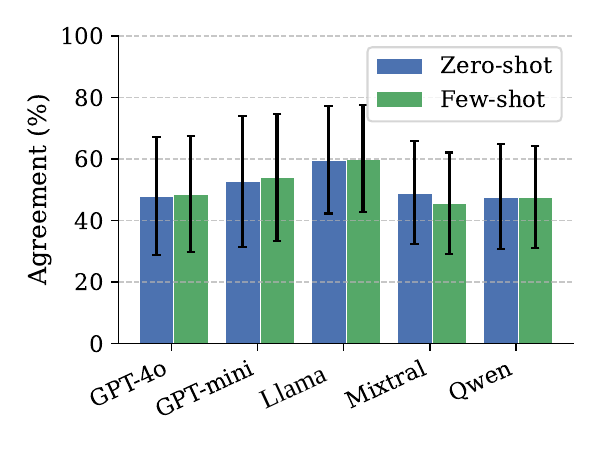}%
        \label{fig:usr_agreement}%
    }
    \subfloat[HelpSteer2 dataset]{%
        \includegraphics[width=0.7\columnwidth]{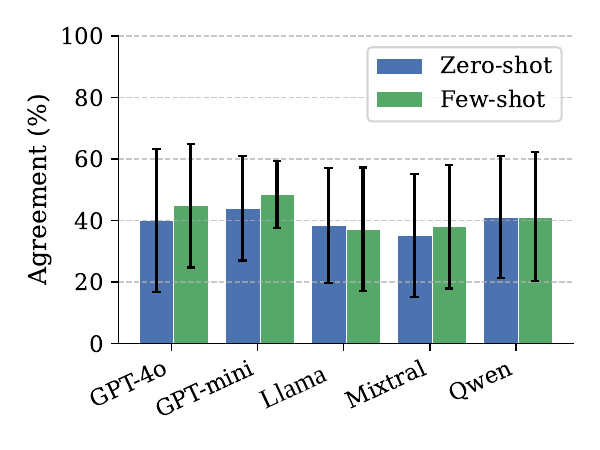}%
        \label{fig:helpsteer2_agreement}%
    }\\
    \subfloat[SummEval dataset]{%
        \includegraphics[width=0.7\columnwidth]{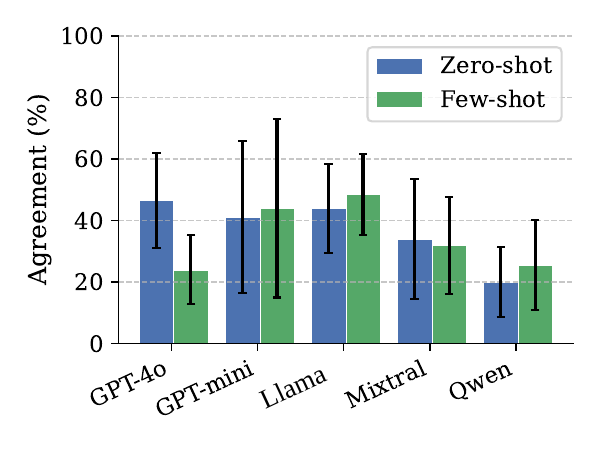}%
        \label{fig:summeval_agreement}%
    }
    \subfloat[SumPubMed dataset]{%
        \includegraphics[width=0.7\columnwidth]{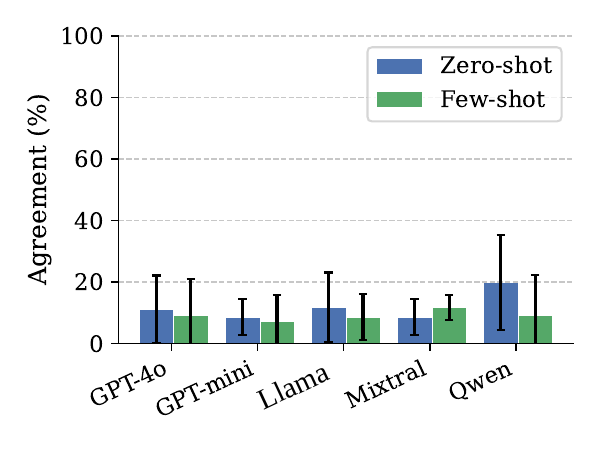}%
        \label{fig:sumpubmed_agreement}%
    }
    \caption{Agreement with 95\% confidence intervals across datasets: (a) USR, (b) HelpSteer2, (c) SummEval, and (d) SumPubMed, between human and LLM scores when using human-defined rubrics.}

    \label{fig:alignment}
\end{figure*}

\subsection{Rubric Application}
\textit{How reliably can LLMs apply self- and human-defined evaluation rubrics when scoring system outputs?}

\begin{table*}[t]
\centering
\small

\begin{tabular}{lcccccccc}
\toprule
\multirow{2}{*}{\textbf{Model}} & \multicolumn{2}{c}{HelpSteer2} & \multicolumn{2}{c}{SummEval} & \multicolumn{2}{c}{SumPubMed} & \multicolumn{2}{c}{USR} \\
\cmidrule(lr){2-3} \cmidrule(lr){4-5} \cmidrule(lr){6-7} \cmidrule(lr){8-9}
  & Agreement (\%) & Corr. & Agreement (\%) & Corr. & Agreement (\%) & Corr. & Agreement (\%) & Corr. \\
\midrule
GPT-4o & 85 & 0.82 & 59 & 0.82 & 92 & 0.53 & 86 & 0.89 \\
GPT-4o-mini & 71 & 0.83 & 67 & 0.83 & 41 & 0.25 & 69 & 0.57 \\
Llama & 63 & 0.62 & 53 & 0.70 & 58 & 0.68 & 72 & 0.58 \\
Mixtral & 79 & 0.95 & 66 & 0.90 & 57 & 0.71 & 74 & 0.62 \\
Qwen & 76 & 0.68 & 72 & 0.82 & 70 & 0.62 & 79 & 0.76 \\
\bottomrule
\end{tabular}%
\caption{Agreement percentage and correlation between zero-shot and few-shot scores per model and dataset (using LLM-generated rubrics).}
\label{tab:llmgen-stability}
\end{table*}

\begin{table*}[!t]
\centering
\small

\begin{tabular}{lcccccccc}
\toprule
\multirow{2}{*}{\textbf{Model}} & \multicolumn{2}{c}{HelpSteer2} & \multicolumn{2}{c}{SummEval} & \multicolumn{2}{c}{SumPubMed} & \multicolumn{2}{c}{USR} \\
\cmidrule(lr){2-3}\cmidrule(lr){4-5}\cmidrule(lr){6-7}\cmidrule(lr){8-9}
  & Agreement (\%) & Corr.  & Agreement (\%) & Corr.  & Agreement (\%) & Corr.  & Agreement (\%) & Corr. \\
\midrule
GPT-4o & 81 & 0.82 & 68 & 0.80 & 69 & 0.73 & 87 & 0.95 \\
GPT-4o-mini & 81 & 0.83 & 68 & 0.82 & 32 & 0.24 & 86 & 0.93 \\
Llama & 76 & 0.72 & 56 & 0.63 & 61 & 0.54 & 86 & 0.93 \\
Mixtral & 75 & 0.81 & 37 & 0.53 & 64 & 0.76 & 80 & 0.89 \\
Qwen & 76 & 0.88 & 73 & 0.77 & 72 & 0.61 & 86 & 0.94 \\
\bottomrule
\end{tabular}%
\caption{Agreement percentage and correlation between zero-shot and few-shot scores per model and dataset (using human rubrics).}
\label{tab:human-stability}
\end{table*}

Models show strong internal consistency when applying their own rubrics. Agreement between zero-shot and few-shot settings (Table~\ref{tab:llmgen-stability}) typically falls between 70–90\%, with correlations above 0.8 for most configurations. GPT-4o and Mixtral are the most robust, maintaining stable rankings across prompts. However, consistency varies by dataset. Scores on HelpSteer2 and USR remain highly correlated, reflecting that conversational tasks with stylistic or pragmatic dimensions are easier to evaluate consistently. In contrast, agreement drops on SummEval and especially SumPubMed, where tasks require factual reasoning or domain knowledge. Lower stability in these datasets indicates that LLMs find it harder to preserve evaluation behavior when rubrics involve technical accuracy or content coverage rather than surface quality.

When using human rubrics, model–human alignment decreases, and variation across domains becomes more pronounced. As shown in Tables \ref{tab:human-stability}, \ref{tab:corr-humanrubric}, and ~\ref{tab:reliability}, correlations with human scores reach $\rho \approx$ 0.8–0.9 for HelpSteer2 and USR, but fall to around 0.5 for SummEval and below 0.3 for SumPubMed. Models capture human preferences for clarity, fluency, and engagement, but struggle with rubrics linked to factuality. Few-shot prompting slightly improves reliability in dialogue-oriented datasets, but has limited or inconsistent benefit for summarization. Figures~\ref{fig:alignment}, \ref{fig:coh_zeroshot_boxviolin}, \ref{fig:spearman_zero}, and \ref{fig:spearman_few} show that agreement variance is largest in the biomedical domain, suggesting that rubric application remains sensitive to task complexity and domain shift.

Comparing across setups, models are far more consistent in reproducing their own scoring patterns than in matching human evaluations. Figure~\ref{fig:coh_zeroshot_boxviolin} examines the \textit{Coherence} rubric, which appears in all datasets, as a common reference. Model and human scores are closely aligned in dialogue data, where coherence reflects local flow and response relevance, but diverge in summarization, where coherence requires document-level structure and logical progression. Wider score dispersion on \textsc{SummEval} and \textsc{SumPubMed} confirms that LLMs' interpretation of coherence differs across prompting conditions and that of humans.

Overall, LLMs apply rubrics consistently across prompting conditions, as reflected by the zero-shot vs.\ few-shot agreement and correlations reported in Tables \ref{tab:llmgen-stability}-\ref{tab:human-stability}, but align less closely with human judgments in tasks that demand domain reasoning or factual consistency. These findings highlight that robustness in rubric use does not yet imply human-level reliability.

\begin{table*}[!t]
\centering
\small

\begin{tabular}{lcccccccc}
\toprule
Model & \multicolumn{2}{c}{HelpSteer2} & \multicolumn{2}{c}{SummEval} & \multicolumn{2}{c}{SumPubMed} & \multicolumn{2}{c}{USR} \\
 \cmidrule(lr){2-3}\cmidrule(lr){4-5}\cmidrule(lr){6-7}\cmidrule(lr){8-9}
 & Zero-shot & Few-shot & Zero-shot & Few-shot & Zero-shot & Few-shot & Zero-shot & Few-shot \\
\midrule
GPT-4o        & 0.49 & 0.56 & 0.55 & 0.58 & 0.14 & 0.30 & 0.81 & 0.82 \\
GPT-4o-mini   & 0.47 & 0.58 & 0.24 & 0.28 & -0.09 & 0.06 & 0.79 & 0.78 \\
Llama     & 0.31 & 0.43 & 0.25 & 0.32 & 0.24 & 0.20 & 0.77 & 0.79 \\
Mixtral & 0.50 & 0.61 & 0.38 & 0.40 & 0.30 & 0.28 & 0.79 & 0.79 \\
Qwen       & 0.65 & 0.63 & 0.46 & 0.45 & 0.19 & 0.28 & 0.81 & 0.80 \\
\bottomrule
\end{tabular}
\caption{Spearman correlation of zero-shot and few-shot LLM scores with human scores across datasets (using human rubrics).}
\label{tab:corr-humanrubric}
\end{table*}

\subsection{Rubric Transfer}
\textit{Do evaluation rubrics generated by one LLM generalize effectively when applied by other models?}

\begin{figure*}[t]
    \centering
    \subfloat[USR: GPT-4o rubric]{%
        \includegraphics[width=0.64\columnwidth]{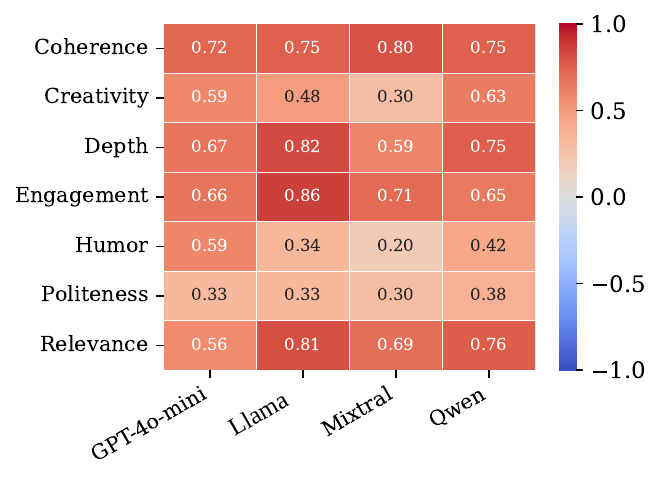}%
        \label{fig:usr_gpt4o_rubric}%
    }
    \subfloat[USR: Llama rubric]{%
        \includegraphics[width=0.64\columnwidth]{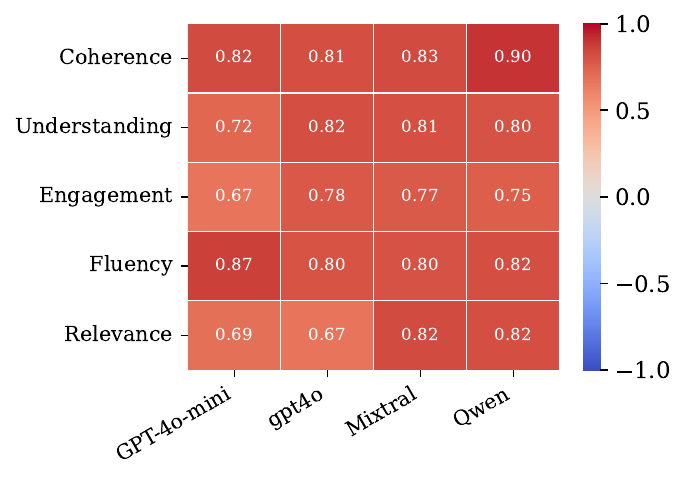}%
        \label{fig:usr_llama_rubric}%
    }\\
    \subfloat[SumPubMed: GPT-4o ]{%
        \includegraphics[width=0.64\columnwidth]{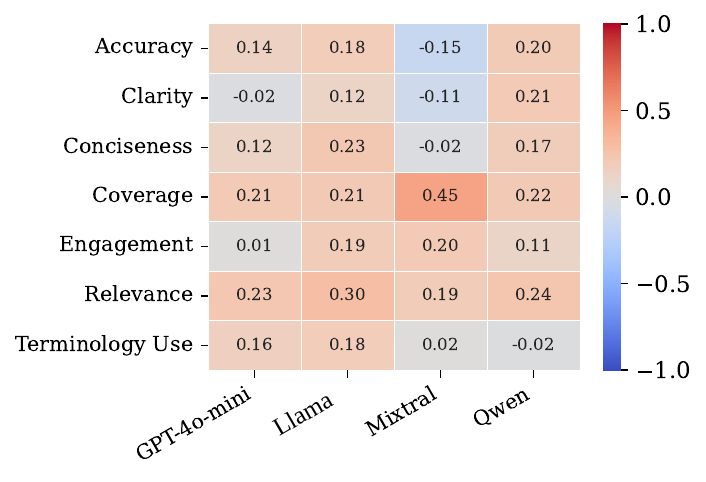}%
        \label{fig:sumpubmed_gpt4o_rubric}%
    }
    \subfloat[SumPubMed: Llama ]{%
        \includegraphics[width=0.64\columnwidth]{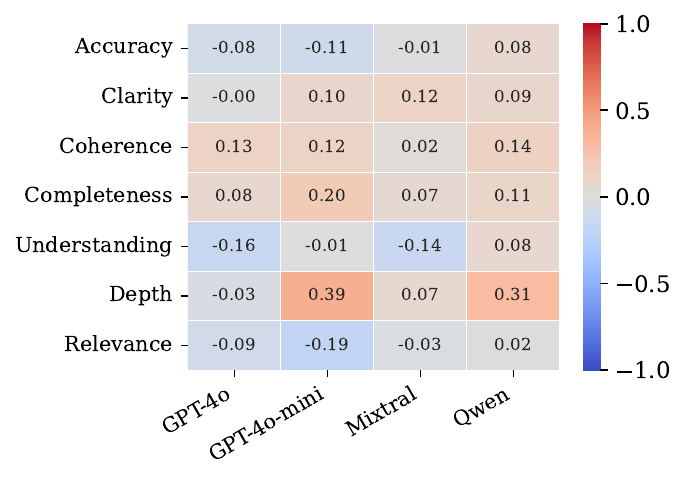}%
        \label{fig:sumpubmed_llama_rubric}%
    }
    \caption{Spearman correlations for two datasets (USR and SumPubMed) and two rubric sources (GPT-4o and Llama).}
    \label{fig:correlation_heatmaps}
\end{figure*}

We investigate whether evaluation rubrics defined by one model can be meaningfully interpreted and applied by others to score the same system outputs.  
This setting captures the extent to which evaluation rubrics are model-specific or reflect shared semantic representations across architectures.  
We analyze two contrasting datasets: \textsc{USR}, an open-domain dialogue benchmark where rubrics emphasize coherence, engagement, and naturalness, and \textsc{SumPubMed}, a biomedical summarization task where scoring depends on factual grounding and terminology precision.  
Rubrics are drawn from GPT-4o and Llama, representing two complementary model families: GPT-4o as a closed, instruction-tuned system and Llama as a widely used open-weight model--allowing us to test cross-family transfer of rubric understanding.

Results across Tables~\ref{tab:usr_llama_reliability}--\ref{tab:sumpubmed_llama_reliability} and Figure~\ref{fig:correlation_heatmaps} reveal clear task-dependent patterns.  
For \textsc{USR}, rubrics generated by both GPT-4o and Llama achieve strong inter-model reliability (ICC $\approx$ 0.7–0.8, $\alpha \approx$ 0.74–0.81) and high cross-model correlations ($\rho \approx$ 0.7–0.9).  
These results indicate that conversational rubrics generalize robustly across models: the relative scoring of system outputs remains consistent even when the rubric originates from a different model.  
Interestingly, while GPT-4o rubrics emphasize dimensions such as \textit{Depth} and \textit{Creativity} and Llama focuses more on \textit{Fluency} and \textit{Contextual Understanding}, both produce comparable consistency across models, suggesting a shared semantic space for dialogue evaluation grounded in pragmatic and linguistic features.  
The strong correlation trends in Figure~\ref{fig:coh_zeroshot_boxviolin} mirror the high reliability observed in the tables, showing that conversational attributes like coherence and engagement are interpreted similarly across models.

In contrast, rubric transfer is substantially weaker for \textsc{SumPubMed}.  
Here, reliability and correlation values fall sharply (ICC < 0.2, $\rho <$ 0.3), and even general rubrics such as \textit{Coherence} and \textit{Relevance} show limited agreement.  
Rubrics tied to factual correctness and terminology—\textit{Accuracy}, \textit{Coverage}, and \textit{Terminology Use}--exhibit near-zero or negative correlations, implying that domain-specific rubrics are not consistently understood across model families.  
The drop in both reliability and correlation highlights that factual reasoning and biomedical knowledge remain model-dependent, and that domain-grounded evaluation rubrics are harder to generalize without shared knowledge representations.

In general, rubric transfer is robust in conversational domains, where evaluation relies on broadly shared linguistic features, but unstable in specialized domains, where scoring depends on factual and terminological accuracy.  
This difference shows that while LLMs converge on surface-level quality judgments, rubric generalization across models still breaks down when evaluation requires domain-specific reasoning or structured content understanding.

\begin{table}[t]
\centering
\footnotesize
\begin{tabular}{p{2.4cm}ccc}
\toprule
Rubric & ICC & $\alpha$ & Fleiss' $\kappa$ \\
\midrule
Coherence                & 0.7906 & 0.7880 & 0.3727 \\
Relevance                & 0.7343 & 0.7311 & 0.3859 \\
Engagement               & 0.7903 & 0.7872 & 0.3858 \\
Contextual Und. & 0.7446 & 0.7415 & 0.3287 \\
Fluency                  & 0.8138 & 0.8108 & 0.4005 \\
\bottomrule
\end{tabular}
\caption{Inter-rater reliability for \textbf{USR} scored with the \textbf{Llama rubric}. Values reported are ICC, Krippendorff’s $\alpha$, and Fleiss’ $\kappa$ across models.}
\label{tab:usr_llama_reliability}
\end{table}

\begin{table}[!t]
\centering
\small
\begin{tabular}{lccc}
\toprule
Rubric & ICC & $\alpha$ & Fleiss' $\kappa$ \\
\midrule
Relevance   & 0.5431 & 0.5332 & 0.4290 \\
Engagement  & 0.6392 & 0.6322 & 0.4513 \\
Coherence   & 0.7305 & 0.7256 & 0.5763 \\
Creativity  & 0.3979 & 0.3931 & 0.3856 \\
Depth       & 0.7810 & 0.7771 & 0.5889 \\
Politeness  & 0.5963 & 0.5847 & 0.3925 \\
Humor       & 0.4929 & 0.4853 & 0.4014 \\
\bottomrule
\end{tabular}
\caption{Inter-rater reliability for \textbf{USR} scored with the \textbf{GPT-4o rubric}. Values reported are ICC, Krippendorff’s $\alpha$, and Fleiss’ $\kappa$ across models.}
\label{tab:usr_gpt4o_reliability}
\end{table}

\begin{table}[t]
\centering
\small
\begin{tabular}{lccc}
\toprule
Rubric & ICC & $\alpha$ & Fleiss' $\kappa$ \\
\midrule
Relevance                & 0.0312 & -0.0379 & -0.0090 \\
Accuracy                 & 0.0723 & 0.0007 & -0.0781 \\
Completeness             & 0.1854 & 0.1570 & 0.0113 \\
Clarity                  & -0.0192 & -0.0352 & -0.0226 \\
Coherence                & 0.0535 & 0.0270 & -0.0007 \\
Depth                    & 0.1323 & 0.0454 & -0.0224 \\
Contextual Und. & 0.0171 & -0.0163 & -0.0635 \\
\bottomrule
\end{tabular}
\caption{Inter-rater reliability for \textbf{SumPubMed} scored with the \textbf{Llama rubric}. Values reported are ICC, Krippendorff’s $\alpha$, and Fleiss’ $\kappa$ across models.}
\label{tab:sumpubmed_llama_reliability}
\end{table}

\begin{table}[!t]
\centering
\small
\begin{tabular}{lccc}
\toprule
Rubric & ICC & $\alpha$ & Fleiss' $\kappa$ \\
\midrule
Relevance        & 0.1942 & 0.1811 & 0.1306 \\
Clarity          & 0.0433 & 0.0432 & 0.0383 \\
Conciseness      & 0.0197 & -0.0296 & -0.0561 \\
Accuracy         & 0.1600 & 0.1125 & 0.0665 \\
Coverage         & 0.0578 & 0.0051 & 0.0122 \\
Terminology Use  & 0.0000 & -0.1054 & -0.1173 \\
Engagement       & 0.1349 & 0.0654 & -0.0534 \\
\bottomrule
\end{tabular}
\caption{Inter-rater reliability for \textbf{SumPubMed} scored with the \textbf{GPT-4o rubric}. Values reported are ICC, Krippendorff’s $\alpha$, and Fleiss’ $\kappa$ across models.}
\label{tab:sumpubmed_gpt4o_reliability}
\end{table}

\section{Discussion}

The results collectively provide two central insights into the capabilities and limits of LLMs as evaluators.  
First, LLMs can reliably generate and apply coherent evaluation rubrics, though their reliability and validity depend on the task domain and model family.  
Second, models interpret their own evaluation rubrics more effectively than they interpret human rubrics or each other's rubrics, revealing insights into how evaluative meaning is represented across systems.

Across tasks, models demonstrate strong internal calibration: they can generate structured rubrics and apply them across prompting conditions.  
Conversational and instruction-following datasets show the highest stability and agreement with human judgments, suggesting that LLMs encode robust linguistic cues for discourse-level quality.  
By contrast, reliability decreases in factual or domain-specific tasks like \textsc{SumPubMed}, where evaluation depends on factual knowledge and terminological precision.  
In these settings, models prioritize fluency and surface coherence over factual correctness, producing scores that are internally consistent but misaligned with human intent.  
This distinction reflects a broader reliability--validity trade-off: \textit{LLMs are procedurally stable but semantically limited}.  
They capture consistent cues embedded in linguistic form and usage, yet their assessments remain shallow when reasoning about facts or content coverage.

Across all analyses, models interpret their own evaluation rubrics most effectively, human rubrics moderately well, and other models' rubrics least consistently.  
When applying self-generated rubrics, they show high cross-prompt agreement, suggesting an internal understanding of their own evaluative constructs. 
Human rubrics yield partial alignment, high for linguistic attributes such as fluency or coherence, but weak for factual or knowledge-oriented rubrics.  
Rubric transfer across models shows the lowest consistency, even when similar rubrics are used, implying that evaluative semantics diverge across architectures.  
Each model therefore develops a distinct ``evaluation dialect,'' shaped by its alignment objectives and training data: internally coherent, yet externally misaligned~\citep{Jones2025UncoveringGI}.

The results suggest that evaluation may itself be a learned linguistic behavior: models can coherently define and apply rubrics yet fail to generalize across architectures, indicating that their notion of quality arises from internal representations rather than fixed external standards. 
In this view, LLMs act not only as evaluators but as agents that help construct the very concept of ``quality'' in generated language.
Their alignment in conversational settings reflects shared communicative priors, whereas their divergence in factual or domain-specific tasks reveals the limits of this emergent consensus.
Future work should build on this capacity to design collaborative evaluation frameworks where human and model rubrics co-evolve, aligning linguistic calibration with factual grounding.
Such approaches would shift evaluation from static benchmarks to dynamic, interpretable, and domain-adaptive practices that reflect both human judgment and model reasoning.

\section{Conclusion}

This paper examined whether LLMs can act as reliable evaluators by generating and applying their own rubrics.  
Using multiple datasets spanning conversational and domain-specific tasks, we analyzed how models define, apply, and transfer evaluative rubrics.  
The results show that LLMs generate coherent rubrics and apply them consistently in open-domain evaluation but struggle with factual and knowledge-intensive tasks.  
Closed-source models such as GPT-4o outperform open-weight models like Llama, generating rubrics that align more closely with human judgments.  
However, models interpret their own evaluation language more effectively than human or peer rubrics, revealing internal coherence but limited shared understanding.  
These findings position LLM-based evaluation as reliable for assessing linguistic quality; however, unreliable for factual reasoning.  
Progress will require shared evaluative representations and hybrid human--LLM frameworks that combine linguistic calibration with domain-specific grounding.

\section*{Limitations}

Our study is limited to a subset of closed-source and open-weight LLMs evaluated on English-language datasets.  
While these models cover diverse families, results may not generalize to smaller, multilingual, or domain-specialized models.  
The analysis focuses on textual outputs and rubric-based evaluations, without examining how multimodal inputs or interactive feedback might alter model behavior.  
Although we include factual and knowledge-intensive tasks, we do not independently verify factual accuracy beyond human reference scores.  
Finally, our work treats LLMs as static evaluators, without modeling temporal drift or adaptation over time.  
Future work should expand to multilingual, multimodal, and dynamic evaluation settings, and investigate how model-specific evaluation dialects evolve with new alignment and fine-tuning strategies.

\section*{Ethical Considerations}

This work uses publicly available datasets and commercially released and open-source models under their respective licenses.  
No personal or sensitive information is introduced, and all evaluations are conducted on benchmarked text.  
We acknowledge potential risks in using LLMs as evaluators, including automation bias, overreliance on model judgments, and reinforcement of model-specific biases.  
By systematically analyzing where such evaluation is reliable and where it remains limited, this study seeks to promote transparency rather than the replacement of human oversight.  
We advocate for hybrid evaluation pipelines in which human judgment remains central, particularly for factual, safety-critical, or socially sensitive domains.  
All data and analysis scripts will be released to support reproducibility and responsible extension of this research.

\subsubsection*{Acknowledgments}
This research is supported in part by the Dutch Science Foundation (NWO) project HAICu with project number NWA.1518.22.105 and in part by the Informatics Institute (IvI) of the University of Amsterdam.

All content represents the opinion of the authors, which is not necessarily shared or endorsed by their respective employers and/or sponsors.

\bibliography{custom,anthology,references}
\clearpage
\appendix
\section{Additional methodology Details}

\subsection{Process Overview}  
The GER-Eval process consists of three steps:  
\begin{enumerate}
    \item Generate a set of rubrics $M$ from $p_\theta$ given $prompt_{gen}(t,\pi)$:  
    \[
    M \sim p_\theta(M \mid prompt_{gen}(t,\pi)).
    \]  

    \item Generate instructions $I(m_i)$ for each rubric $m_i \in M$:  
    \[
    I(m_i) \sim p_\theta(I \mid m_i, t).
    \]  

    \item Apply each rubric with instructions $I(m_i)$ to outputs $Y$, producing reasoning and ratings where: $(r_{i,j}, s_{i,j}) \sim p_\theta(r, s \mid prompt_{eval}(t, m_i, I(m_i), y_j))$

\end{enumerate}  

The final output is a structured representation of rubrics and their associated scores:  
\[
\{(m_i, I(m_i), \{(y_j, r_{i,j}, s_{i,j})\}_{j=1}^N )\}_{i=1}^k.
\]

\subsection{Datasets}
\label{app:datasets}
\begin{enumerate}
    \item \textbf{SummEval} \cite{fabbri-etal-2021-summeval} is a news domain dataset, where the task is abstractive summarization. The outputs are system-generated summaries of CNN/Daily Mail articles, with human ratings provided on coherence, consistency, fluency, and relevance.  

\item  \textbf{HelpSteer2} \cite{wang2024helpsteer2} represents the instruction-following domain, where the task is to generate helpful responses to user prompts. The outputs are model-generated chat responses, annotated by humans on five attributes of helpfulness.  

\item  \textbf{USR} \cite{mehri-eskenazi-2020-usr} is an open-domain dialogue dataset, where the task is dialogue response generation. The outputs are continuations of dialogue contexts, and expert evaluators provide ratings across six dimensions of dialogue quality.  

\item \textbf{SumPubMed} \cite{gupta-etal-2021-sumpubmed} is situated in the biomedical research domain, where the task is long-form scientific summarization. The outputs are system-generated summaries of PubMed articles, with expert judgments on readability, coherence, redundancy, and informativeness.  

\end{enumerate}

Together, these datasets cover news, conversational, instructional, and scientific domains, offering a broad set of text generation tasks on which to test the generalizability of GER-Eval.

\subsection{Models}
\label{app:models}

We evaluate a set of closed- and open-source LLMs commonly used in recent LLM-as-judge and automated evaluation studies, covering different model scales and training regimes. Our selection includes GPT-4o and GPT-4o-mini (OpenAI), Mixtral-8$\times$22B, Llama-3.3-70B, and Qwen2.5-72B, enabling analysis across proprietary and open-weight model families.

\begin{itemize}
  \item \textbf{GPT-4o}~\citep{gpt4-report} serves as a high-capacity proprietary judge model, frequently adopted as a reference in evaluation pipelines, with a \textbf{128K} context window.

  \item \textbf{GPT-4o-mini}~\citep{gpt4-report} is a lower-cost, reduced-capacity variant used to study how evaluation reliability changes under constrained model capacity.

  \item \textbf{Mixtral 8$\times$22B}~\citep{jiang2024mixtralexperts} represents open-weight mixture-of-experts models, allowing us to examine rubric generation and application under sparse activation settings (\textbf{64K} context).

  \item \textbf{Qwen\,2.5-72B-Instruct}~\citep{qwen2} is a strong open-weight dense model with long-context support (\textbf{128K}), commonly used in recent evaluation and benchmarking work.

  \item \textbf{Llama\,3.3-70B-Instruct}~\citep{Touvron2023-llama} serves as a widely adopted open-weight baseline for instruction-following and judgment tasks, supporting long-context evaluation.
\end{itemize}

\subsubsection{Model Configurations}  
All models were run with temperature set to 0 for scoring tasks to ensure determinism, using greedy decoding. For rubric generation, we used temperature = 0.7 with nucleus sampling ($p=0.9$) to encourage diversity in rubric generation. The maximum output length was set to 512 tokens for rubric generation and 256 tokens for scoring.  

For GPT-4o and GPT-4o-mini, prompts were formatted using the system–user message structure provided by the OpenAI API. For Mixtral-8x22B, prompts were serialized as plain text and accessed via the Mistral API. For Llama-3.3-70B and Qwen2.5-72B, prompts were likewise serialized into plain text and accessed through the DeepInfra API. In all cases, we normalized the prompt content to ensure consistency across APIs. Role descriptions, when included, were applied uniformly across models.  

Outputs were requested in structured JSON containing fields for rubric name, description, scale, instruction, reasoning, and score. All outputs were automatically validated; malformed responses were re-prompted up to three times before being discarded. Random seeds were fixed across runs for reproducibility. All experiments were conducted via API inference only, and no local GPU compute was used.

\subsubsection{Implementation details} 
We experimented with 50 data points from each dataset across all the conditions.
For each dataset, we randomly sampled three contexts per task description. In the contrastive condition, each context was paired with one positive and one negative response based on dataset-specific quality signals (e.g., overall scores in USR, helpfulness in HelpSteer2, relevance in SummEval, and informativeness/overlap/focus in SumPubMed).  

During scoring, when categorical labels were available in the dataset (e.g., Helpsteer), the model was prompted to produce categorical judgments; otherwise, ratings were mapped to a continuous 1–5 scale.

\subsubsection{Rubric Statistics: Breadth, Unique, and Alignment}
\label{app:align}
\paragraph{Breadth.}
We define Breadth as the total number of criteria produced by the model for a given setting.

\paragraph{Unique.}
To account for redundant or paraphrased criteria, we compute Unique by de-duplicating the generated set using semantic similarity over criterion descriptions. We embed each description with \texttt{jinaai/jina-embeddings-v3} and merge criteria whose cosine similarity exceeds a fixed threshold ($\tau=0.82$), yielding a de-duplicated set of criteria. We manually inspected a set of merged and non-merged pairs to verify that high-similarity pairs correspond to genuine semantic duplicates (e.g., \textit{Brevity} vs.\ \textit{Conciseness}), and used the embedding-based merging thereafter.

\paragraph{Alignment (Align.\%).}
We estimate overlap with the dataset's human rubric via LLM-based tagging. We use GPT-4o (temperature $0$) as a tagger conditioned on the task description, the human rubric items (teacher), and the generated criteria (student). For each generated criterion, the tagger assigns one or more human rubric item labels when the teacher description is substantially covered; otherwise it assigns a novel ``other aspect'' label. A criterion is counted as unseen if none of its tags correspond to a human rubric item. We report:
\[
\text{Align.(\%)} = 100 \times \left(1 - \frac{\#\text{Unseen}}{\#\text{Unique}}\right).
\]

\subsubsection{Prompts used for generation and scoring}
\label{app:prompts_for_generation}
Below we show sample prompts used in rubric generation~\ref{lst:metric-generation-prompt}, rubric scoring instructions~\ref {lst:instruction-prompt}, and rubric scoring~\ref{lst:scoring-prompt}.

\begin{lstlisting}[caption={Rubric Generation Prompt}, label={lst:metric-generation-prompt}]
METRIC_GENERATION_PROMPT = ```
{Task description}
{samples_prompt}
Based on this information, define a set of metrics that should be used to assess the quality of this task.
For each metric, only include: Name, Description, Scale
'''
\end{lstlisting}

\begin{lstlisting}[caption={Instruction Prompt}, label={lst:instruction-prompt}]
INSTRUCTION_PROMPT = ``` Please generate scoring instruction for this {Metric} and {Description}.  
Evaluation Criteria: {Metric} Evaluation Steps: '''
\end{lstlisting}

\begin{lstlisting}[caption={Scoring Prompt}, label={lst:scoring-prompt}]
SCORING_PROMPT = ```
 {Description} Please score the following output based on the: 
 Evaluation Criteria: {Metric}
 Evaluation Steps: {Steps}
 {Additional_info}
 Sample of Evaluation: {Sample}
 Evaluation of {Metric_name}:
 '''
\end{lstlisting}

\section{End-to-End Example of GER-Eval}
\label{app:example}
Table~\ref{tab:end_to_end_example} provides an end-to-end example of rubric generation and application for HelpSteer2. We show excerpts for one generated criterion for brevity.

\begin{table*}[!t]
\small
\centering
\setlength{\tabcolsep}{6pt}
\begin{tabular}{p{0.18\textwidth} p{0.78\textwidth}}
\toprule
\textbf{Stage} & \textbf{Excerpt (HelpSteer2, GPT-4o, Task-only)} \\
\midrule
Rubric generation prompt & \textit{``Define clear and independent metrics to evaluate the quality of assistant responses. For each metric, output only: Name, Description, Scale.''} \\
\midrule
Generated rubric & \textbf{Name:} Creativity. \textbf{Description:} Assesses the originality of the response and whether it introduces novel ideas or perspectives. \textbf{Scale:} 1--5 (1 = not creative, 5 = highly creative). \\
\midrule
Generated scoring instruction & \textit{``Read the user prompt and the response. Judge whether the response contributes unique or inventive ideas. Assign a score from 1 to 5. Provide one sentence explaining the score.''} \\
\midrule
Scoring prompt (constructed by GER-Eval) & \textit{``Using the rubric \textbf{Creativity} with the instruction above, evaluate the following response on a 1--5 scale and provide a brief justification.''} \\
\midrule
Model output & \textbf{Score:} 4. \textbf{Justification:} \textit{``The response introduces several novel narrative elements that originally extend beyond the prompt.''} \\
\bottomrule
\end{tabular}
\caption{End-to-end example of GER-Eval for one generated rubric (excerpts shown for brevity).}
\label{tab:end_to_end_example}
\end{table*}

\section{Supplemental Results}

\begin{figure}[!t]
    \centering
    \includegraphics[width=\columnwidth]{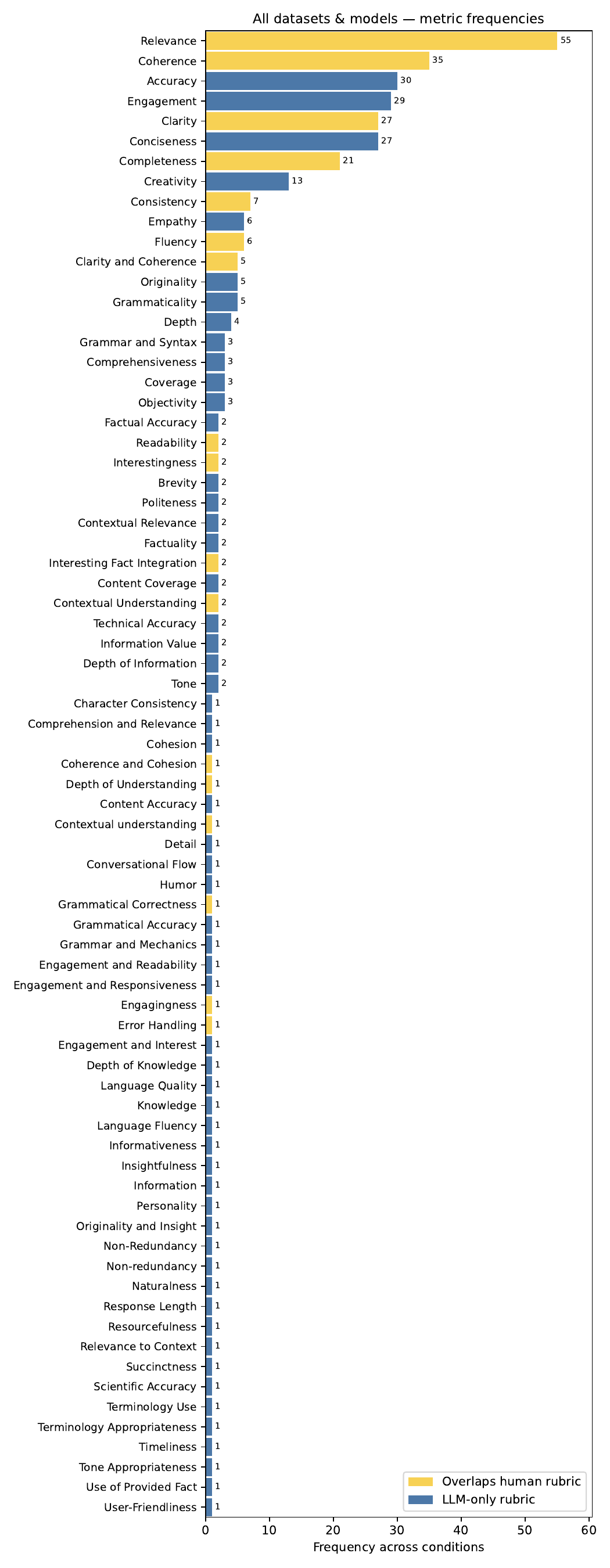}
    \caption{Rubric Frequencies from all datasets and conditions and all models, also overlapping with human rubrics.}
    \label{fig:GLOBAL__metric_frequencies}
\end{figure}

This appendix provides additional analyses that complement the main results in the paper. We first report model- and dataset-specific task-sensitive rubric names generated under the few-shot setting (Table~\ref{tab:task-specific}), alongside the corresponding human-defined task-specific rubrics for each dataset (Table~\ref{tab:gt_ts_names}). We then present the global frequency of generated rubrics across all models, datasets, and prompting conditions, including overlap with human rubrics (Figure~\ref{fig:GLOBAL__metric_frequencies}). 

\begin{figure}[!t]
    \centering
   
    \begin{subfigure}{0.48\columnwidth}
        \includegraphics[width=\linewidth]{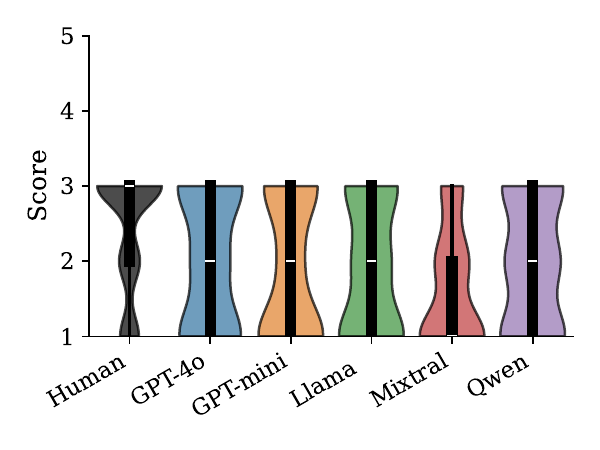}
        \caption{USR}
        \label{fig:coh_usr_zero}
    \end{subfigure}
    \hfill
    \begin{subfigure}{0.48\columnwidth}
        \includegraphics[width=\linewidth]{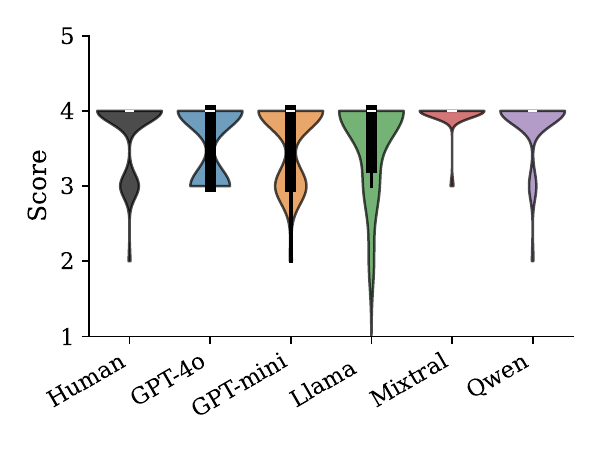}
        \caption{HelpSteer2}
        \label{fig:coh_helpsteer_zero}
    \end{subfigure}
    
    \par\medskip 
    
    \begin{subfigure}{0.48\columnwidth}
        \includegraphics[width=\linewidth]{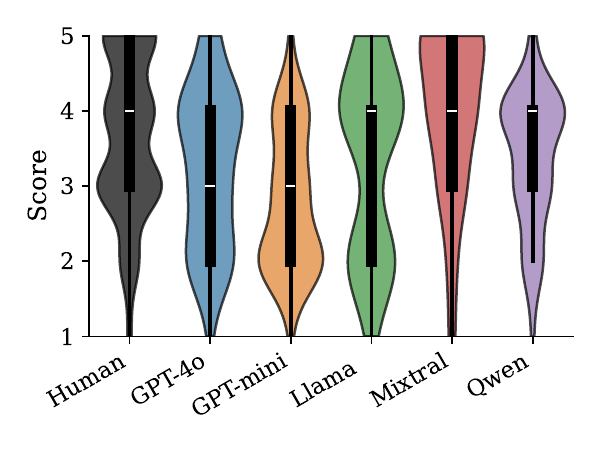}
        \caption{SummEval}
        \label{fig:coh_summeval_zero}
    \end{subfigure}
    \hfill
    \begin{subfigure}{0.48\columnwidth}
        \includegraphics[width=\linewidth]{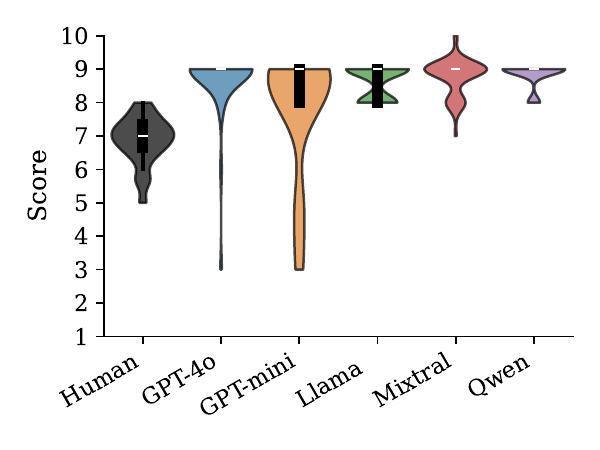}
        \caption{SumPubMed}
        \label{fig:coh_sumpubmed_zero}
    \end{subfigure}

    \caption{Score distributions for the \textbf{Coherence} rubric under the Zero-shot setting across datasets. 
    Each violin shows the score distribution, with a box inside indicating the median and quartiles.}
    \label{fig:coh_zeroshot_boxviolin}
\end{figure}

\begin{figure*}[!t]
    \centering
    \subfloat[USR]{%
        \includegraphics[width=0.485\textwidth]{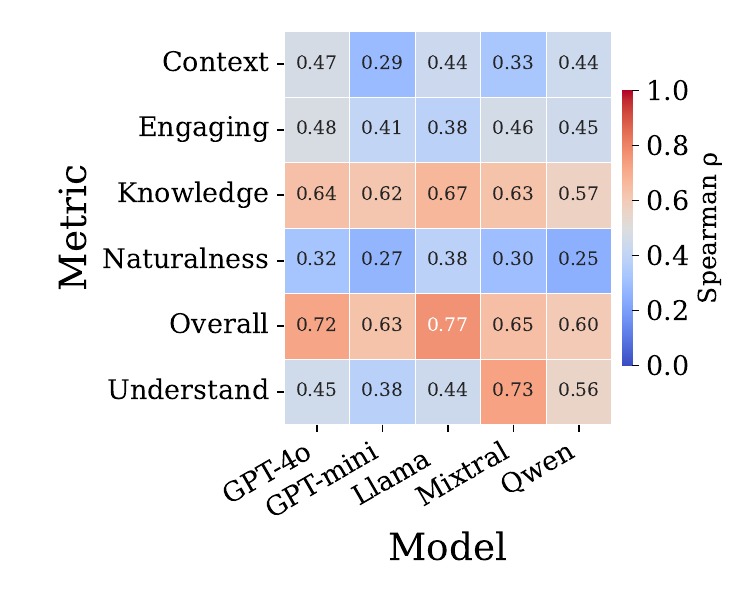}%
        \label{fig:spearman_usr_zero}%
    }\hfill
    \subfloat[HelpSteer2]{%
        \includegraphics[width=0.485\textwidth]{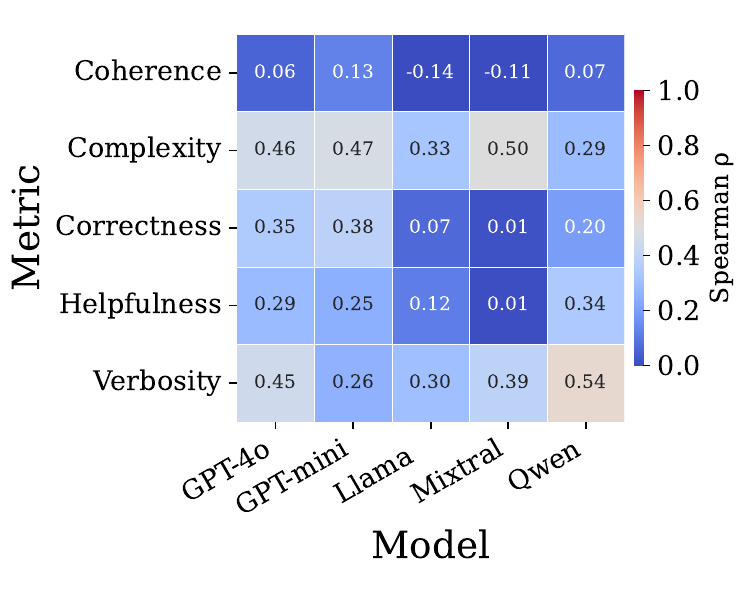}%
        \label{fig:spearman_helpsteer_zero}%
    }\\
    \subfloat[SummEval]{%
        \includegraphics[width=0.485\textwidth]{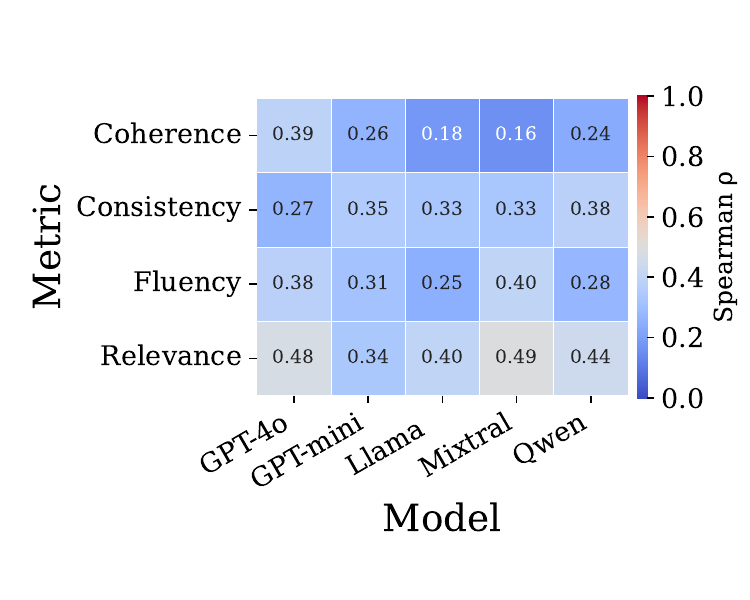}%
        \label{fig:spearman_summeval_zero}%
    }\hfill
    \subfloat[SumPubMed]{%
        \includegraphics[width=0.485\textwidth]{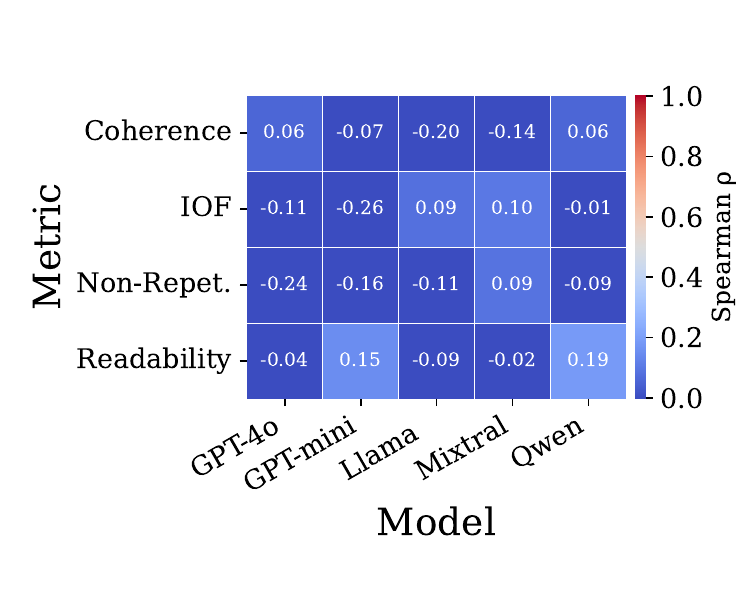}%
        \label{fig:spearman_sumpubmed_zero}%
    }
    \caption{Spearman correlation ($\rho$) between LLM and human scores under the \textbf{Zero-shot} condition across four datasets.}
    \label{fig:spearman_zero}
    \vspace{-1mm}
\end{figure*}

\begin{figure*}[!t]
    \centering
    \subfloat[USR]{%
        \includegraphics[width=0.485\textwidth]{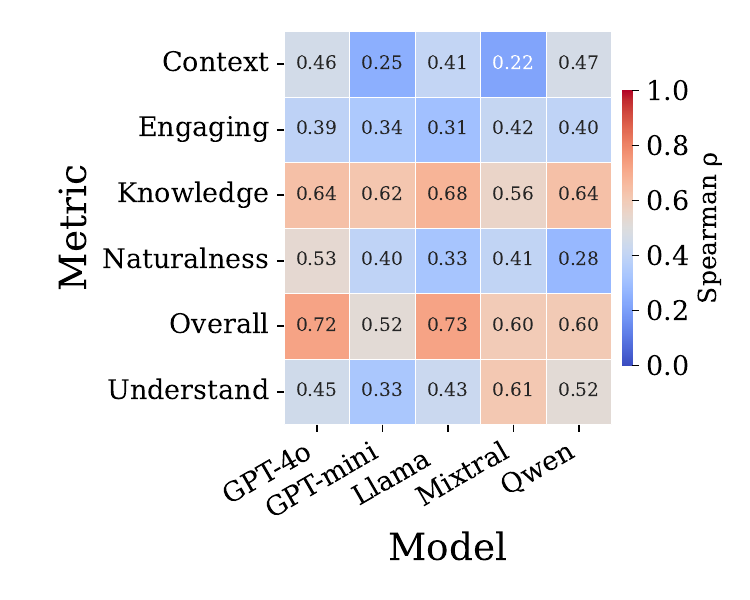}%
        \label{fig:spearman_usr_few}%
    }\hfill
    \subfloat[HelpSteer2]{%
        \includegraphics[width=0.485\textwidth]{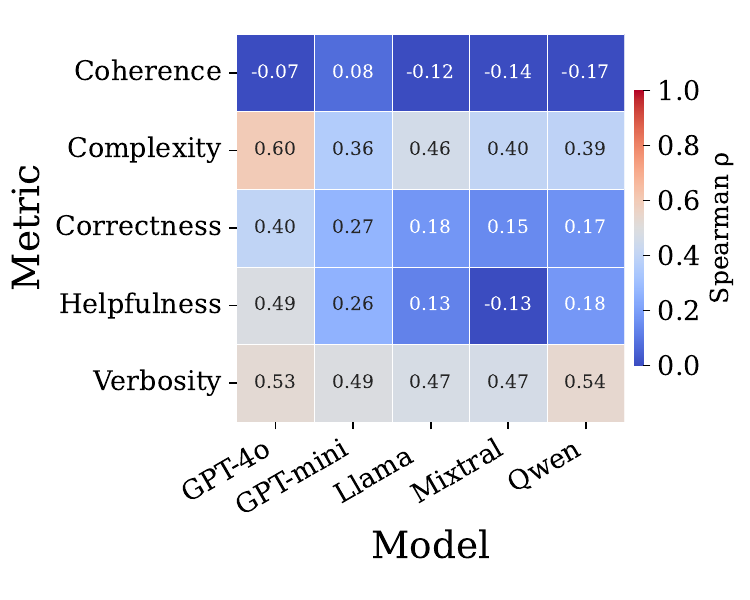}%
        \label{fig:spearman_helpsteer_few}%
    }\\
    \subfloat[SummEval]{%
        \includegraphics[width=0.485\textwidth]{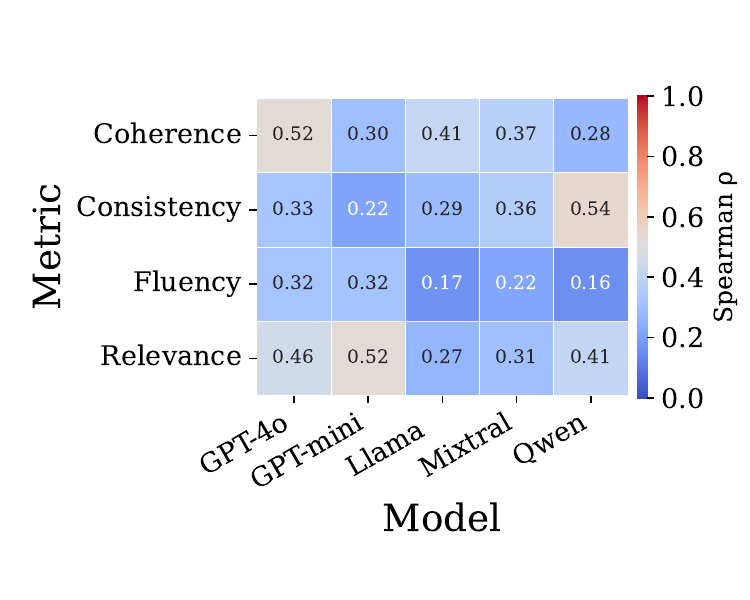}%
        \label{fig:spearman_summeval_few}%
    }\hfill
    \subfloat[SumPubMed]{%
        \includegraphics[width=0.485\textwidth]{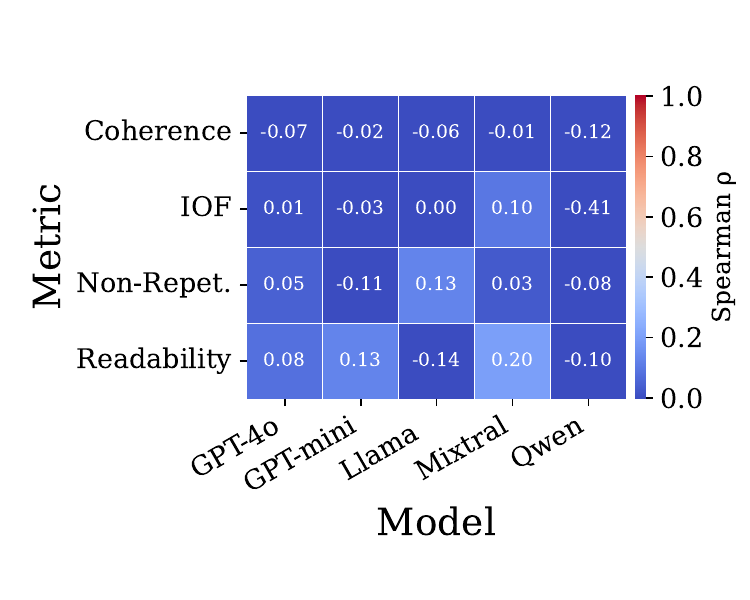}%
        \label{fig:spearman_sumpubmed_few}%
    }
    \caption{Spearman correlation ($\rho$) between LLM and human scores under the \textbf{Few-shot} condition across four datasets.}
    \label{fig:spearman_few}
    \vspace{-2mm}
\end{figure*}

\begin{figure*}[!t]
    \centering
    \subfloat[HelpSteer2]{%
        \includegraphics[width=0.4\textwidth]{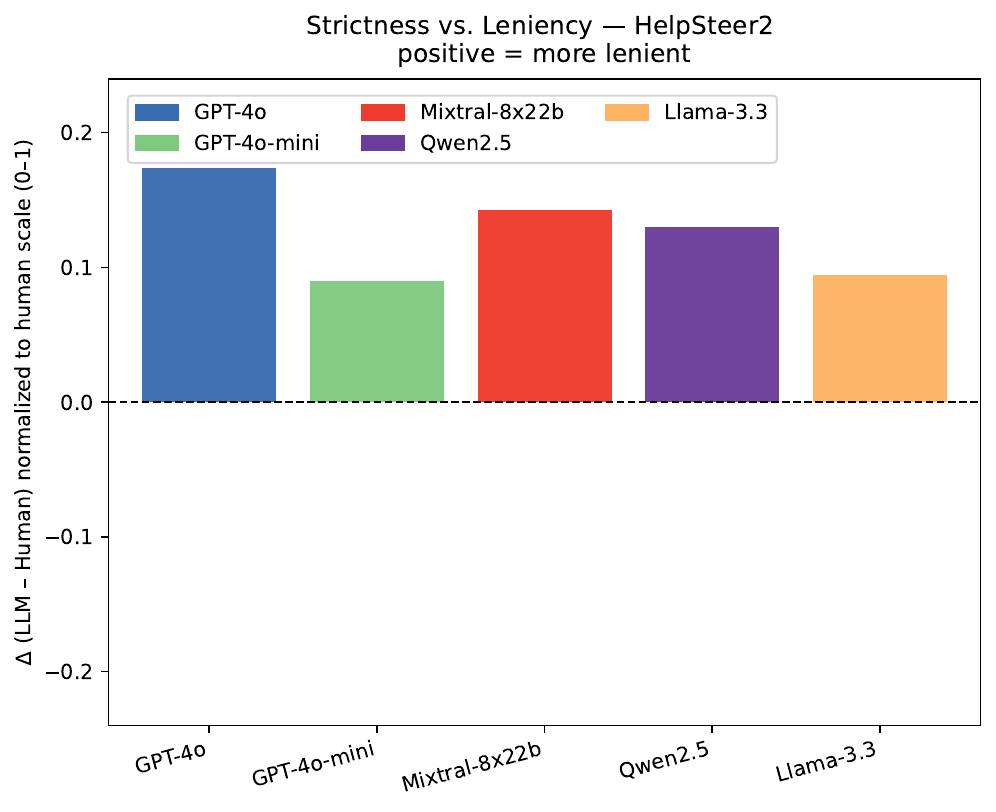}%
        \label{fig:strict_helpsteer2_zero}%
    }\hfill
    \subfloat[SumPubMed]{%
        \includegraphics[width=0.4\textwidth]{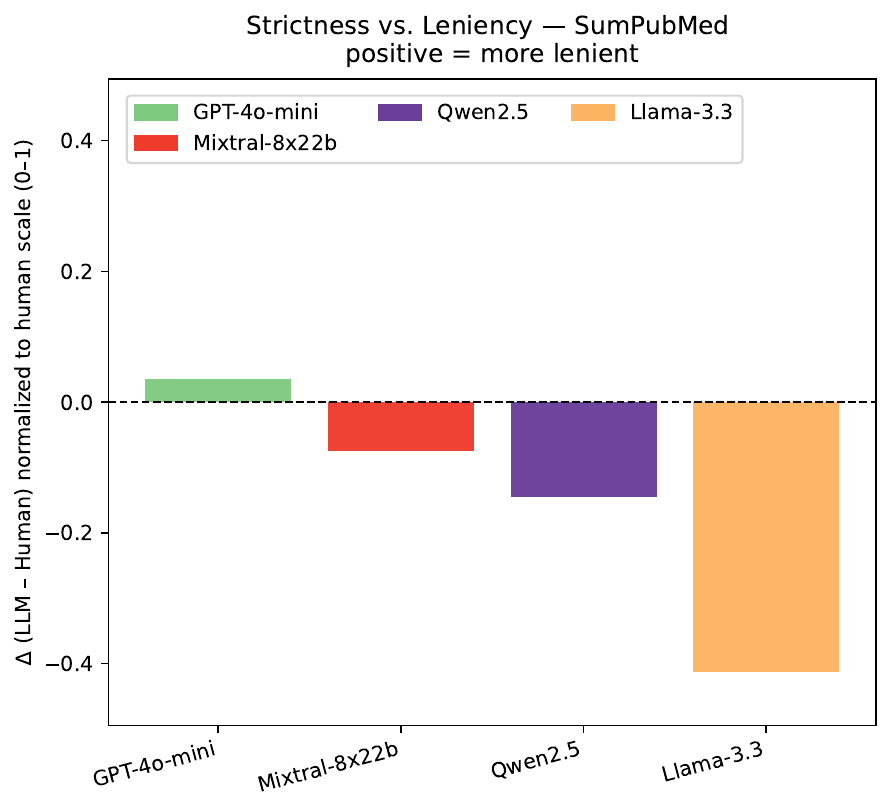}%
        \label{fig:strict_sumpubmed_zero}%
    }\\
    \subfloat[SummEval]{%
        \includegraphics[width=0.4\textwidth]{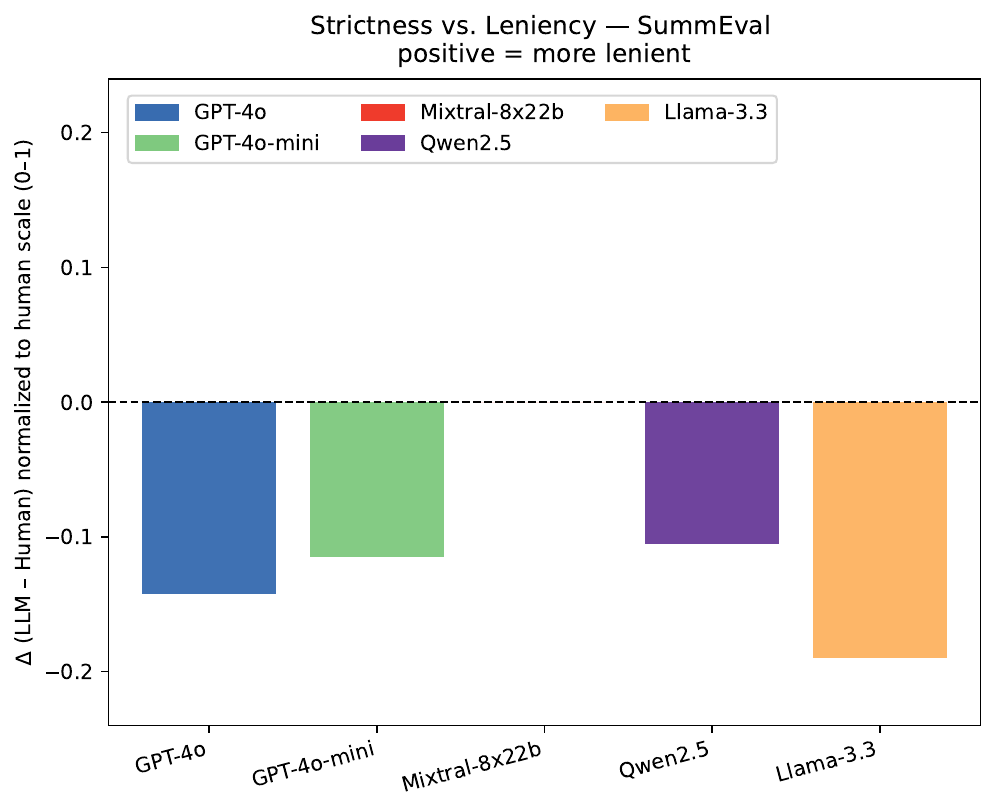}%
        \label{fig:strict_summeval_zero}%
    }\hfill
    \subfloat[USR]{%
        \includegraphics[width=0.4\textwidth]{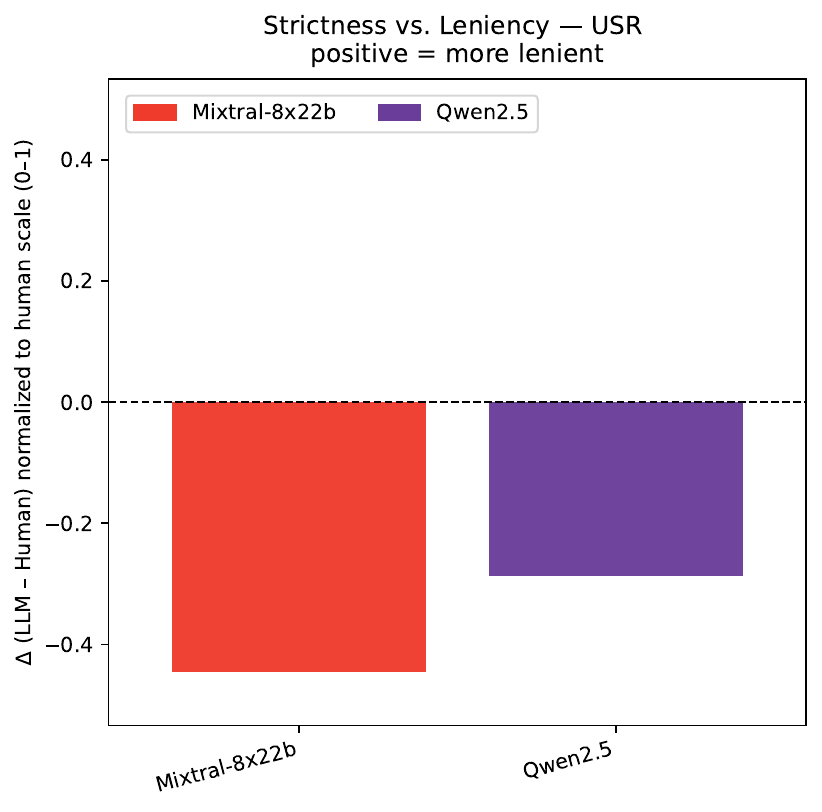}%
        \label{fig:strict_usr_zero}%
    }
    \caption{Strictness versus leniency of LLM scoring behavior across datasets under the zero-shot setting.}
    \label{fig:strict_leniency}
\end{figure*}

To further characterize evaluation behavior, we report inter-rater reliability statistics under both zero-shot and few-shot conditions using human-defined rubrics (Table~\ref{tab:reliability}). We additionally visualize score distributions for the \emph{Coherence} rubric across datasets in the zero-shot setting (Figure~\ref{fig:coh_zeroshot_boxviolin}). Finally, we include correlation analyses between LLM and human scores under zero-shot and few-shot conditions (Figures~\ref{fig:spearman_zero} and~\ref{fig:spearman_few}), as well as dataset-level analyses of strictness versus leniency in LLM scoring behavior (Figure~\ref{fig:strict_leniency}).

\begin{table*}[!t]
\centering
\small

\resizebox{\textwidth}{!}{%
\begin{tabular}{llcc}
\toprule
\textbf{Dataset} & \textbf{Model} & \textbf{Task-specific rubrics}  \\
\midrule
HelpSteer2 & GPT-4o & Tone \\
HelpSteer2 & GPT-4o-mini & Depth of Knowledge, Error Handling  \\
HelpSteer2 & Mixtral & Engagement, Originality  \\
HelpSteer2 & Qwen & Clarity and Coherence, Consistency, Depth of Information, Engagement and Responsiveness, Relevance to Context, Technical Accuracy, User-Friendliness  \\
HelpSteer2 & Llama & Conversational Flow, Empathy  \\
SummEval & GPT-4o & Clarity and Coherence, Content Coverage, Language Quality  \\
SummEval & GPT-4o-mini & Accuracy, Completeness, Objectivity  \\
SummEval & Mixtral & --  \\
SummEval & Qwen & --  \\
SummEval & Llama & Readability  \\
SumPubMed & GPT-4o & Coverage, Terminology Use  \\
SumPubMed & GPT-4o-mini & Clarity and Coherence, Content Coverage, Engagement and Readability, Technical Accuracy  \\
SumPubMed & Mixtral & Clarity, Objectivity  \\
SumPubMed & Qwen & Comprehensiveness, Objectivity  \\
SumPubMed & Llama & Clarity, Depth  \\
USR & GPT-4o & Creativity, Humor, Politeness  \\
USR & GPT-4o-mini & Coherence, Grammar and Syntax  \\
USR & Mixtral & Character Consistency, Coherence and Cohesion, Contextual Relevance, Engagement and Interest, Grammatical Accuracy, Interesting Fact Integration  \\
USR & Qwen & Creativity, Empathy, Knowledge  \\
USR & Llama & Engagement  \\
\bottomrule
\end{tabular}
}
\caption{Task sensitivity: rubrics names per model and dataset (With Examples Setting (Few-shot)).}
\label{tab:task-specific}
\end{table*}

\begin{table*}[t]
\centering
\small
\caption{Ground-truth(Human) \emph{task-specific} rubrics.}
\label{tab:gt_ts_names}
\resizebox{\textwidth}{!}{%
\begin{tabular}{lp{0.82\textwidth}}
\toprule
\textbf{Dataset} & \textbf{Task-specific rubrics} \\
\midrule
HelpSteer2 & Clarity, Completeness, Complex Language, Correctness, Helpfulness, Safe, Simple, Succinct, Understanding, Unsafe, Verbose Language \\
SummEval & Consistency, Fluency, Relevance \\
SumPubMed & Informativeness, Overlap and Focus, Non-Repetition and no factual Redundancy, Readability \\
USR & Interesting, Maintains Context, Natural, Understandable, Uses Knowledge \\
\bottomrule
\end{tabular}
}
\end{table*}

\begin{table*}[!t]
\centering
\small
\begin{tabular}{llcccccc}
\toprule
Dataset & Rubric 
& \multicolumn{3}{c}{Zero-shot} 
& \multicolumn{3}{c}{Few-shot} \\
\cmidrule(lr){3-5} \cmidrule(lr){6-8}
 &  & ICC(2) & Krippendorff $\alpha$ & Fleiss $\kappa$ 
    & ICC(2) & Krippendorff $\alpha$ & Fleiss $\kappa$ \\
\midrule
\multirow{5}{*}{HelpSteer2}
  & Helpfulness & 0.683 & 0.679 & 0.347 & 0.526 & 0.521 & 0.352 \\
  & Correctness & 0.452 & 0.448 & 0.299 & 0.366 & 0.362 & 0.224 \\
  & Coherence   & 0.311 & 0.297 & 0.293 & 0.209 & 0.180 & 0.163 \\
  & Complexity  & 0.609 & 0.598 & 0.256 & 0.585 & 0.574 & 0.219 \\
  & Verbosity   & 0.688 & 0.680 & 0.330 & 0.646 & 0.637 & 0.282 \\
\midrule
\multirow{4}{*}{SummEval}
  & Relevance   & 0.463 & 0.440 & 0.172 & 0.390 & 0.377 & 0.244 \\
  & Consistency & 0.616 & 0.603 & 0.166 & 0.537 & 0.521 & 0.148 \\
  & Fluency     & 0.328 & 0.279 & 0.084 & 0.309 & 0.280 & 0.068 \\
  & Coherence   & 0.527 & 0.509 & 0.181 & 0.446 & 0.434 & 0.175 \\
\midrule
\multirow{4}{*}{SumPubMed}
  & Non-Re      & 0.123 & 0.014  & -0.011 & 0.164 & 0.130  & 0.060 \\
  & Coh         & 0.156 & 0.130  &  0.052 & -0.010 & -0.034 & -0.076 \\
  & Read        & 0.095 & -0.002 &  0.021 & 0.028 & -0.105 & -0.071 \\
  & IOF         & 0.312 & 0.263  &  0.038 & 0.134 & 0.075  & 0.006 \\ %
\midrule
\multirow{6}{*}{USR} 
  & Understandable    & 0.637 & 0.630 & 0.629 & 0.608 & 0.599 & 0.597 \\
  & Natural           & 0.554 & 0.547 & 0.404 & 0.666 & 0.660 & 0.473 \\
  & Maintains Context & 0.652 & 0.646 & 0.475 & 0.577 & 0.566 & 0.373 \\
  & Engaging          & 0.657 & 0.651 & 0.539 & 0.637 & 0.631 & 0.448 \\
  & Uses Knowledge    & 0.747 & 0.743 & 0.742 & 0.754 & 0.749 & 0.748 \\
  & Overall           & 0.717 & 0.712 & 0.236 & 0.759 & 0.754 & 0.314 \\
\bottomrule
\end{tabular}
\caption{Inter-rater reliability across LLMs under \textbf{Zero-shot} and \textbf{Few-shot} conditions. Values show ICC(2), Krippendorff’s $\alpha$, and Fleiss’s $\kappa$ for each evaluation rubric across all four datasets using human rubric.}
\label{tab:reliability}
\end{table*}

\end{document}